\documentclass{article}
\PassOptionsToPackage{numbers, sort&compress}{natbib}
\usepackage[preprint]{neurips_2026}

\usepackage[utf8]{inputenc} 
\usepackage[T1]{fontenc}    
\usepackage{hyperref}       
\usepackage{url}            
\usepackage{booktabs}       
\usepackage{amsfonts}       
\usepackage{nicefrac}       
\usepackage{microtype}      
\usepackage[table]{xcolor}         

\usepackage{amsmath}
\usepackage{amssymb}
\usepackage[pdftex]{graphicx}
\graphicspath{{./Figures/}}

\title{PINNs Failure Modes are Overfitting}

\author{%
    Nigel T. Andersen\textsuperscript{1} \quad
    Takashi Matsubara\textsuperscript{1,2} \\
    \textsuperscript{1}Graduate School of Information Science and Technology \\
    Hokkaido University \\
    Sapporo, Japan \\
    \textsuperscript{2}RIKEN Center for Advanced Intelligence Project (AIP) \\
    Tokyo, Japan \\
    \texttt{\{n.andersen, matsubara\}@ist.hokudai.ac.jp} \\
}

\begin{document}

\maketitle

\begin{abstract}

Physics-Informed Neural Networks (PINNs) are a common class of machine learning-based partial differential equation (PDE) solvers which train a network to represent a solution by minimizing a residual loss that encodes the PDE. Despite their successes, they are known to fail on certain simple equations, converging to an incorrect solution despite low loss. These failure modes have garnered significant attention in the literature over the past several years, motivating both architectural and optimization based solutions. By directly visualizing the residual, we show that failure modes are the result of overfitting: the loss is minimized on the collocation points, but not elsewhere.
Applying regularization causes the failure modes to vanish.
Finally, we extend double backpropagation over the full set of residuals, and use it to achieve state-of-the-art performance on four standard failure mode equations with up to $23\times$ fewer collocation points and a vanilla architecture.

\end{abstract}

\section{Introduction}

Using neural networks to represent the solution of partial differential equations has garnered much attention over the past several years. Many different approaches have emerged, from methods using networks as one component of the solution \cite{ psichogios1992hybrid, han2018solving}, to those based on classical methods like the Ritz and Galerkin methods \cite{lagaris1998artificial, yu2018deep}. Other approaches have found success in approximating the solution operators instead of the solutions themselves \cite{li2020neural,li2020fourier,chen1995universal,li2024physics,rossi2005functional,lu2021learning}. However, the formalism of Physics-Informed Neural Networks (PINNs) \cite{raissi2019physics} remains one of the most studied. PINNs use a neural network to directly approximate the solution of the partial differential equation, and train it by minimizing the residual of the PDE operator applied to the network in tandem with losses representing the boundary and initial conditions. The architecture is simple to implement, and has shown great success.

Yet despite this, PINNs can struggle on relatively simple problems. Most notably is the convection / linear advection equation, which acts to translate the initial condition. While PINNs learn the solution easily for weak convection speeds, they collapse for strong convection, learning instead the trivial zero solution over most of the domain. This convergence to the incorrect solution is commonly referred to as a failure mode~\cite{krishnapriyan2021characterizing}.

Failure modes have proven an immense challenge for PINNs, and various diagnoses and fixes have been proposed. Reasons for failure have been broadly attributed to spectral bias and loss imbalances between the domain and boundary terms~\cite{wang2022and,liu2022unified}, propagation issues from the initial condition into the domain~\cite{wu2025propinn,daw2023mitigating,wang2024respecting}, as well as the complexity of the loss landscape~\cite{rathore2024challenges,liu2024preconditioning,xu2025fp64}.
Proposed solutions include loss reweighting, adaptive sampling techniques, novel optimization methods, new architectures, and notably, a simple upgrade from floating point 32 (FP32) to floating point 64 (FP64)~\cite{wang2022and,liu2022unified,daw2023mitigating,wang2024respecting,rathore2024challenges,liu2024preconditioning,xu2025fp64}.

\begin{figure}
    \centering
    \includegraphics[width=\linewidth]{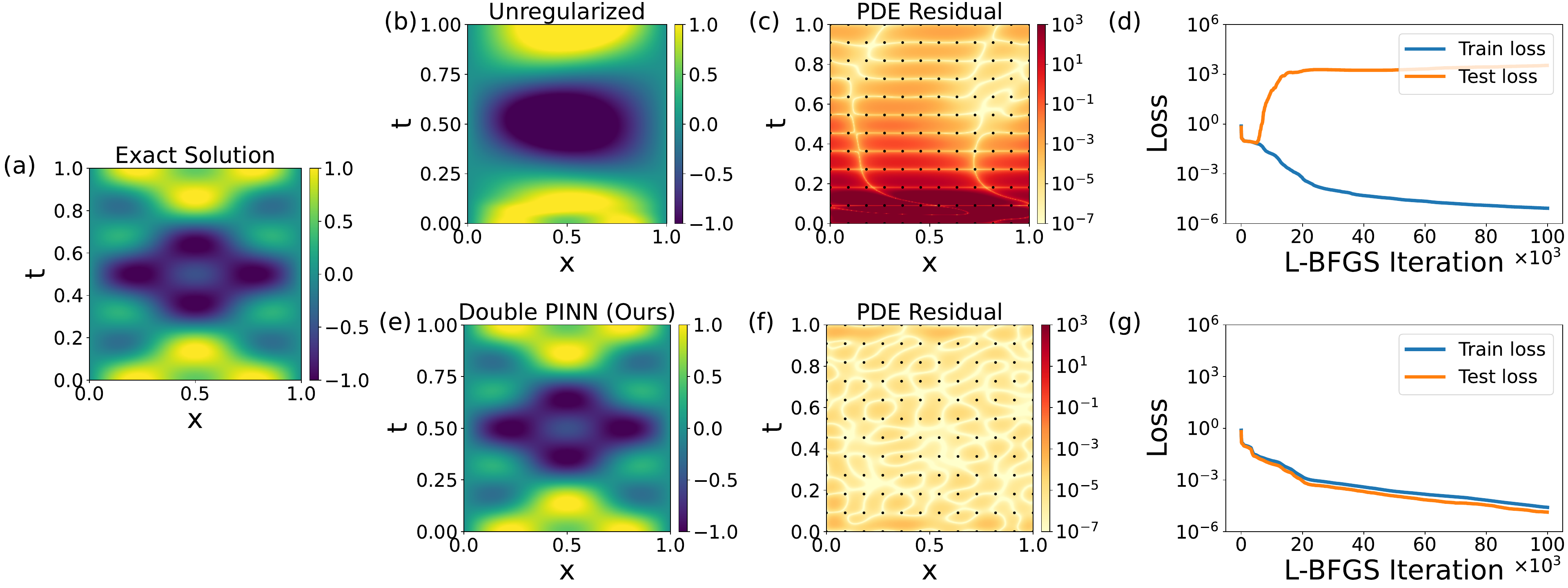}
    \caption{Exact solution to the wave equaiton (a), compared to the PINNs failure mode (b). Unregularized PINNs (b)-(d) minimize the residual only at the collocation points, resulting in failure to learn the solution even though the objective loss has been minimized. Regularizing with double backpropagation eliminates this, resulting in successful training of vanilla PINNs (e)-(g).}
    \label{fig:convec}
\end{figure}

While the focus on the optimization difficulties has delivered many useful architectures, and improved our understanding of PINNs and machine learning optimization more generally, even the state-of-the-art method (FP64 optimization)~\cite{xu2025fp64}, still struggles with a relatively high error, leading to a suboptimal performance when compared to traditional methods. Interestingly, as we will show, the high-precision method can still suffer from failure modes.

If a robust optimization procedure cannot fully eliminate the collapse of PINNs into failure modes, then what is the underlying cause inhibiting convergence to the true solution? We argue that there are two fundamental issues, that have been conflated under the broad umbrella of optimization difficulties. The first is the genuine optimization problems that have been diagnosed in detail~\cite{rathore2024challenges,liu2024preconditioning,xu2025fp64}. These inhibit convergence in higher order problems, such as the wave or Allen-Cahn equations. Previous literature on failure modes has mostly targeted this perspective.

The second, which has remained undiagnosed, is overfitting. By directly visualizing the PDE residual of a failure mode, we show that in regions of high residual the loss has been minimized only at the collocation points, leading to a system well converged to the wrong solution. This diagnosis is consistent across equations, and is further confirmed by a divergence between a training and test loss when the system enters a failure mode.

We then demonstrate that the success of increased precision relies on using a dense collocation grid. Reducing the number of points causes failure even at higher precision, implying that FP64 training in part relies on over-determining the problem. This too is consistent with overfitting: more collocation points makes it harder for the network to memorize the data. Yet increasing the points directly increases the cost of training. 

By incorporating standard $L^2$ regularization used for reducing overfitting in broader machine learning, we show the failure modes vanish even at FP32, while also allowing us to use fewer collocation points over the domain ($<1000$, compared to $>10000$ in Refs~\cite{zhao2024pinnsformer, xu2025sub, xu2025fp64}). Then, we generalize double backpropagation~\cite{drucker1992improving} to PINNs, applying it to each of the residual terms, which both improves training efficiency and final accuracy.

Combining this double-backpropagation gradient regularization with FP64 optimization, then allows us to outperform state-of-the-art by over an order of magnitude in error on the full benchmark suite of failure mode equations, while using far fewer points.

\paragraph{Contributions.} Our main contributions are as follows:

\begin{itemize}

\item We reveal that while PINNs failure modes are considered to be caused by the optimization landscape, they are in fact caused by overfitting, where the optimization minimizes the loss only at the collocation points in regions of high residual. This changes the understanding of why and when PINNs fail.

\item Using this understanding, we show that adding a standard $L^2$ regularizer takes the optimization process from failure to success, even at FP32.

\item Finally, we extend double backpropagation (gradient regularization) to PINNs, which fits in naturally into the PINNs framework, mitigates failure modes, improves optimization efficiency and reduces collocation requirements. This allows us to achieve state-of-the-art performance on the standard set of failure mode equations with a vanilla architecture.

\end{itemize}

\section{Related Works}

\paragraph{Physics Informed Neural Networks and Their Failure Modes}

A large body of literature has focused on applying machine learning methods to the solution of PDEs~\cite{sirignano2018dgm, han2017deep, yu2018deep, lagaris1998artificial, dissanayake1994neural, han2018solving}. PINNs are one of the most popular formulations, showing how the methods can be applied to both forward and inverse problems, as well as the data-driven discovery of PDEs themselves~\cite{raissi2019physics}. Our focus here is on the forward problem, meaning using the neural network itself as an approximation to the solution. This is in contrast to the distinct but related field of neural operators, which learn a solution functional~\cite{li2020neural,li2020fourier,li2024physics,chen1995universal,lu2021learning}.

Despite their promise and success in high-dimensional problems, PINNs have been known to exhibit optimization difficulties, collapsing and failing to learn the solution for relatively simple low dimensional problems. These have been termed failure modes, and result in a seemingly successful optimization, with a loss near zero, yet having learned a solution far from the desired result~\cite{krishnapriyan2021characterizing, wang2022and}. These failure modes have been extensively studied, and several techniques and architectures have been proposed with the goal of ameliorating them~\cite{krishnapriyan2021characterizing, wang2022and, wang2021eigenvector, mcclenny2023self, wang2024respecting, xu2025sub, daw2023mitigating, liu2024preconditioning, lau2024pinnacle, moro2025solving, wu2025propinn, liu2025bwler, hwang2024dual}. Notably, it has been recently proposed that the optimization difficulties stem from insufficient numerical precision inhibiting optimization, rather than from successful optimization to an incorrect solution~\cite{xu2025fp64}. We will show that while optimization can be improved with precision, the origin of the failure modes lies in overfitting, not optimization difficulty.

\paragraph{Regularization}

Regularization is a standard tool in deep learning~\cite{goodfellow2016deep,murphy2022probabilistic}. Its goal is to improve the generalizability of the machine learning model by reducing overfitting to the training dataset. Common versions include applying a soft constraint on the weights in the form of an addition to the loss, such as with $L^2$ and $L^1$ style regularizations~\cite{krogh1991simple,tibshirani1996regression}, 
and derivative methods like double backpropagation and Sobolev training~\cite{drucker1992improving,czarnecki2017sobolev}. Regularization has rarely been applied to PINNs. Occasionally, the PINN losses have been referred to as regularizers themselves, particularly when data is also being fit~\cite{wang2021understanding, raissi2019physics}. However, this is a different meaning from how we use the terminology: without the existing data, they are simply the PINNs objectives, rather than a tool used to reduce overfitting, which is the sense with which we employ the term "regularizer" here.
Gradient regularization has been applied to PINNs in a style similar to double backpropagation, though only on the domain objective~\cite{yu2022gradient}. Sobolev training was considered in Ref.~\cite{son2023sobolev}. While gradient regularized PINNs are simple and improve performance, they kept the theoretical analysis of why as an open question~\cite{yu2022gradient}. Sobolev PINNs are powerful, but require separate structure on a PDE by PDE basis~\cite{son2023sobolev}. Neither were applied to PINNs failure modes.

\section{Experimental Setup} \label{sec:exp}

This section describes the problem formulation of PINNs, explaining how the network is trained to predict the solution to the PDE. 
Our network architecture itself follows closely to the original PINNs formulation, and uses 4 hidden layers, each with 128 neurons and $\tanh$ activation functions. The network is initialized using the Glorot normal initialization~\cite{glorot2010understanding}, and training was done using the Limited Memory BFGS (L-BFGS) algorithm~\cite{liu1989limited}, which is also standard~\cite{raissi2019physics,krishnapriyan2021characterizing}. Matching current failure modes literature, we use a grid collocation method over the domain, and linearly spaced points on the boundaries and initial conditions~\cite{xu2025fp64}. Experiments using FP32 were run on a Nvidia Titan RTX card, while experiments with FP64 precision were run on a Nvidia A100 GPU. In addition to the training collocation set, a much larger set of random collocation points over the domain is used as a test set, in addition to denser boundaries and initial conditions. (see Appendix~\ref{app:exp} for details). We wrote our implementation using the JAX library~\cite{jax2018github}.

\begin{figure}[t]
    \centering
    \includegraphics[width=\linewidth]{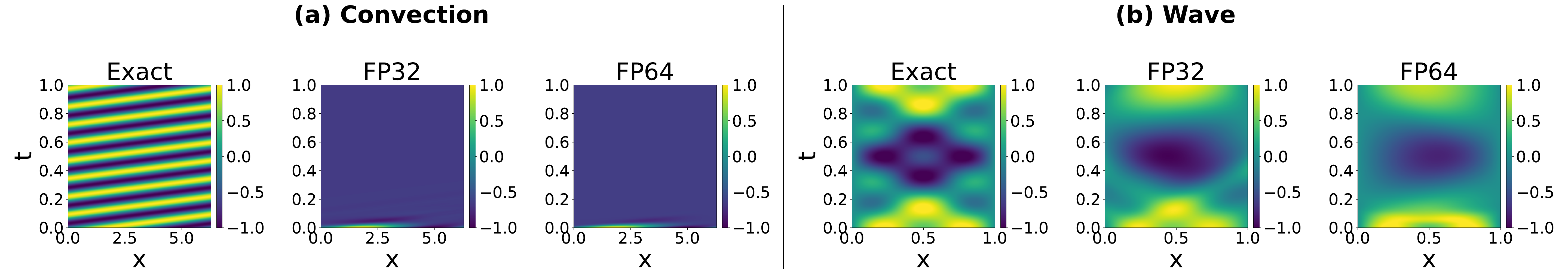}
    \caption{Examples of failures modes at (a) FP32 (convection equation), and (b) FP64 (wave equation). Increasing numerical precision is not sufficient (or necessary) to avoid failure modes.}
\end{figure}

\subsection{Physics Informed Neural Networks}

PINNs aim to train a deep neural network $u_\theta$, parametrized by $\theta$, to approximate the solution to a PDE. The PDE is represented by a differential operator $\mathcal{F}$ acting on the true solution $u$, which for the time dependent problems considered here gives
\begin{equation}
    \mathcal{F}(u(x, t)) = 0, \quad x \in \Omega \subset \mathbb{R}^d, \quad t \in [0,T]
\end{equation}
where $x$ is the $d$-dimensional spatial variable representing the input over the domain $\Omega$, and $t$ the 1d temporal variable. The network learns to approximate $u(x,t)$ by $u_{\theta}(x,t)$ by minimizing the loss $\mathcal{L}(u_\theta)$, which consists of a domain loss $\mathcal{L}_{\mathcal{F}}(u_\theta)$, set of boundary losses $\{\mathcal{L}_{\mathcal{B}i}(u_\theta)\}$, and set of initial condition losses $\{\mathcal{L}_{\mathcal{I}}(u_\theta)\}$:
\begin{equation}
    \min_{\theta} \mathcal{L}(u_\theta)
    = \lambda_\mathcal{F} \mathcal{L}_\mathcal{F}(u_\theta)
    + \sum_j^{N_\mathcal{B}} \lambda_{\mathcal{B}j} \mathcal{L}_{\mathcal{B}j}(u_\theta)
    + \sum_j^{N_\mathcal{I}} \lambda_{\mathcal{I}j} \mathcal{L}_{\mathcal{I}j}(u_\theta)
\end{equation}
where $N_\mathcal{B}$, $N_\mathcal{I}$, are the number of boundary and initial conditions, which depend on the PDE of interest, and
\begin{gather}
    \mathcal{L}_{\mathcal{F}}(u_\theta)  = \frac{1}{n_d} \sum_{k=1}^{n_d} \| \mathcal{F}(u_\theta(x^k_d,t^k_d)) \|^2, \\
    \mathcal{L}_{\mathcal{B}}(u_\theta)  = \frac{1}{n_b} \sum_{k=1}^{n_b} \| \mathcal{B}(u_\theta(x^k_b,t^k_b)) \|^2, \quad
    \mathcal{L}_{\mathcal{I}}(u_\theta)  = \frac{1}{n_i} \sum_{k=1}^{n_i} \| \mathcal{I}(u_\theta(x^k_i,t^k_i)) \|^2
\end{gather}
are the domain, boundary, and initial losses, with $n_d$, $n_b$, and $n_i$ referring to the number of collocation points sampled at domain, boundaries, and initial conditions. Where multiple boundary or initial conditions are required, we use the same number of collocation points for each of them. We follow the original PINNs formulation and set $\lambda_\mathcal{F} = \lambda_\mathcal{B} = \lambda_\mathcal{I} = 1$~\cite{raissi2019physics}.

If a subset of data representing the real solution $u$ is known, the above loss values are sometimes known as physics-based regularizers, as they enforce the PDE prior knowledge during optimization in addition to an extra loss term minimizing the error between the network and known solution points $\| u(x_i,t_i) - u_\theta(x_i,t_i)\|^2$~\cite{raissi2019physics,wang2021understanding}. 
This is true while known data is present. However, as is common in modern PINNs, we work without prior knowledge or data of the solution, and the loss terms serve to enforce the correct domain, boundary, and initial conditions that the solution requires in order to satisfy the PDE. When discussing regularization later on, we use it in the sense of improving the performance of the model, rather than referring to the role of the core loss terms themselves.

Throughout this paper we use a regularizer we call double PINN, our extension of double backpropagation~\cite{drucker1992improving} to the full set of PINN losses:
\begin{equation} \label{eq:double_pinn}
    \mathcal{L}(u_\theta)
    =  \mathcal{L}_\mathcal{F}(u_\theta)
    + \frac{\lambda_r}{2} \| \nabla_{x,t} \mathcal{F} \|^2
    + \sum_j^{N_\mathcal{B}} \left(\mathcal{L}_{\mathcal{B}_j}(u_\theta)
    + \frac{\lambda_r}{2} \| \nabla_{t} \mathcal{B}_j \|^2 \right)
    + \sum_j^{N_\mathcal{I}} \left( \mathcal{L}_{\mathcal{I}_j}(u_\theta)
    + \frac{\lambda_r}{2} \| \nabla_{x} {\mathcal{I}_j} \|^2 \right)
\end{equation}
Double-backpropagation regularizes the gradient of the loss with respect to the network inputs, and double PINN extends this to each PINN loss. By flattening the residuals, double PINN penalizes the network for reducing the losses at the collocation points at the expense of the surrounding region. We define it here for reference; the motivation is developed in Sec.~\ref{sec:opt_v_fail} and a comparison to alternatives in Sec.~\ref{sec:reg}.

\section{Failure Modes are Caused by Overfitting} \label{sec:overfitting}

\begin{figure}[t]
    \centering
    \includegraphics[width=\linewidth]{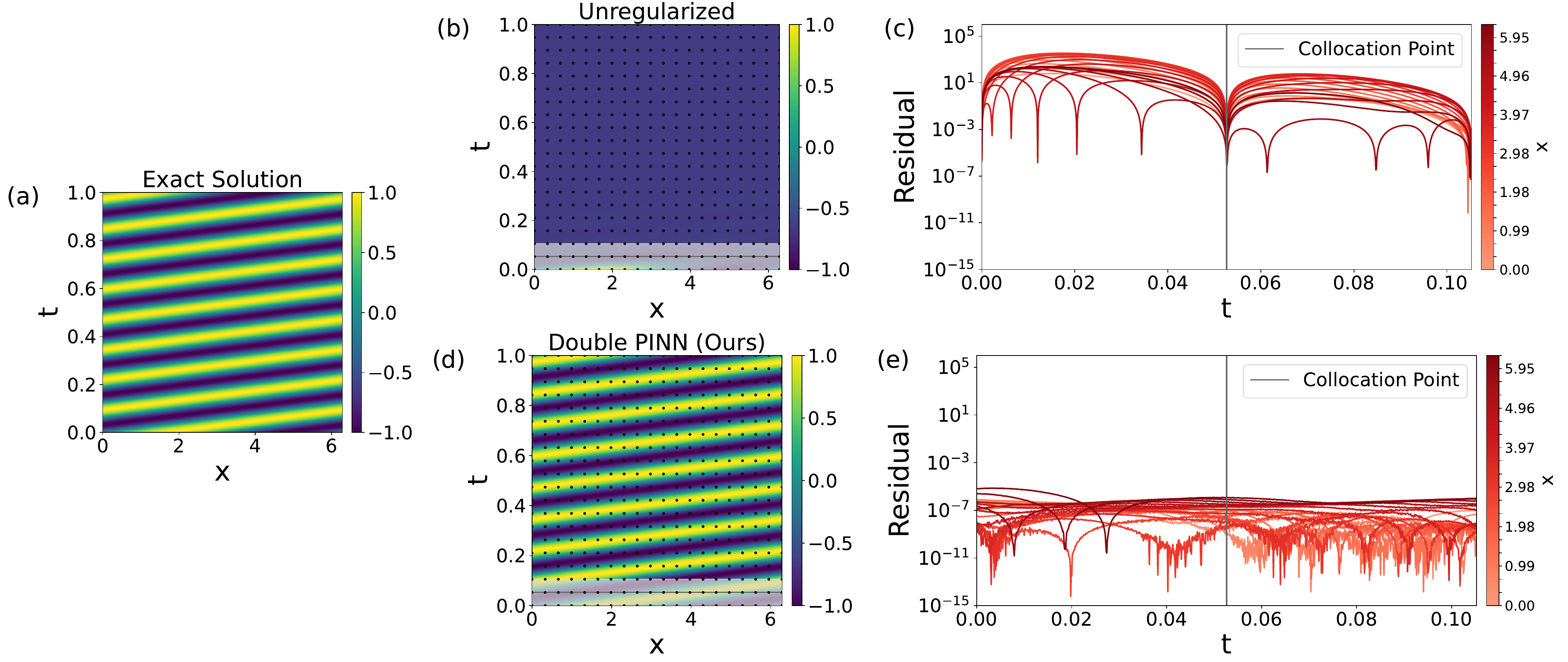}
    \caption{Failure and success for the convection equation at FP32. Exact solution is shown in (a), while the unregularized version is shown in (b), and the regularized double PINN in (d). The domain residual over the highlighted segment in (b) and (d) is plotted for vanilla PINNs in (c), and double PINN in (e). Each curve is the domain residual as a function of $t$ for a given $x$ value. The collocation points are indicated by the vertical line. Unregularized PINN show a clear preference for reducing the residual at the collocation point, while double PINN remains smooth and shows no such preference.}
    \label{fig:wave}
\end{figure}

There are two leading explanations for the root cause of the failure modes.
The first, is that there is an insurmountable loss-barrier, which traps the optimizer in a local minimum with high loss~\cite{krishnapriyan2021characterizing}. Recently a second proposal emerged, suggesting that instead of a local minimum the issue is unsuccessful optimization due to limited precision inhibiting navigation of the loss landscape during optimization~\cite{xu2025fp64}.

In this section, we challenge both interpretations. By examining the residual as well as the loss during training, we show that the network has successfully been optimized to a global minimum with low loss, not a high loss basin, but that this has occurred by overfitting to the collocation points used during training. We then show that an upgrade to FP64 does not rescue the optimization.

\subsection{Optimization Difficulty or Failure Mode} \label{sec:opt_v_fail}

We consider two classic failure mode equations: the convection equation, and the wave equation. Both are relatively simple equations, standard in failure mode literature, and have known closed form solutions.

\subsubsection{Convection Equation (FP32 Failure)}

First we start with the 1D convection equation with periodic boundary conditions, which acts to transport the initial condition $u(x,t=0)$, with advection speed $\beta$. The problem is notable, because it is first order, has a simple solution, and can be readily solved by PINNs for small $\beta$, while failing if $\beta$ is made large. The exact problem formulation can be found in App.~\ref{app:convection}.
For the advection speed $\beta$, a common choice is $\beta =50$, which we use here.

To examine the loss basin hypothesis, we first consider a typical training run that results in a failure mode. Figure~\ref{fig:convec}(a) shows the expected exact solution, while Fig.~\ref{fig:convec}(b) shows the result of the PINNs training. This is the prototypical failure mode, with the PINN outputting zero over most of the domain. 
The loss-basin hypothesis asserts that the PINNs is stuck in a local minimum, and that a true global minimum exists elsewhere~\cite{krishnapriyan2021characterizing}.
Yet looking at the loss in Fig.~\ref{fig:convec}(d) shows smooth optimization to $\sim 10^{-1}$, indicating successful optimization to a near-global minimum. It is the test loss, rather than the training loss, that remains large.
While we have not seen a direct comparison between a test and train loss in the PINNs literature before, we consider this divergence the hallmark of a PINNs failure mode, and it is something that was consistently observed across all of our tests. 

Fig.~\ref{fig:convec}(c) further shows that even though the residual is large in the boundary region, it has been minimized at the collocation points. Rather than a local minimum, this indicates overfitting, where the PINN has learned the dataset (the training collocation points) at the expense of generality (satisfying the solution everywhere). The prescription is then regularization.

The same system with added regularizer is shown in Fig.~\ref{fig:convec} (e)-(g), where we no longer see overfitting at the collocation points, and the test and train loss decrease in tandem.

\subsubsection{Wave Equation (64 bit failure)} \label{sec:wave}

What about the same-basin hypothesis, where the failure mode is suspected to be caused by insufficient numerical precision~\cite{xu2025fp64}? We consider this by examining the wave equation at both FP32 and FP64. The wave equation describes the time evolution of waves, and is detailed in App.~\ref{app:wave}

Reference~\cite{xu2025fp64} reported successful optimization on the wave equation using a $101\times101=10201$ grid over the domain. A large amount of points makes it difficult for the network to memorize the data. If our hypothesis of overfitting is correct, we should expect reduced performance, and failure, as the number of collocation points is reduced, even at FP64.

Figure~\ref{fig:wave} shows FP64 PINNs trained on the wave equation with a reduced $12\times 12=144$ points. While successful on the $10201$ grid, FP64 PINNs fail on the $144$ grid. However, regularizing the same training run allows the optimization to succeed (Fig.~\ref{fig:wave}(d)), even on such a small grid. The PDE residual loss plotted against a time slice of the collocation points is shown in (c) and (e), with the grey lines indicating their position. The unregularized PINN has notable minima exactly at the collocation points. The regularized version shows no such correlation between collocation and residual minima.

\subsection{Regularization Combined with Precision is Optimal}

Section~\ref{sec:wave} showed that precision is not sufficient to avoid failure, and that problems with a low amount of data remain susceptible to failure modes caused by overfitting. Yet increasing the precision can still be helpful, especially for stiff equations.

\begin{figure}
    \centering
    \includegraphics[width=\linewidth]{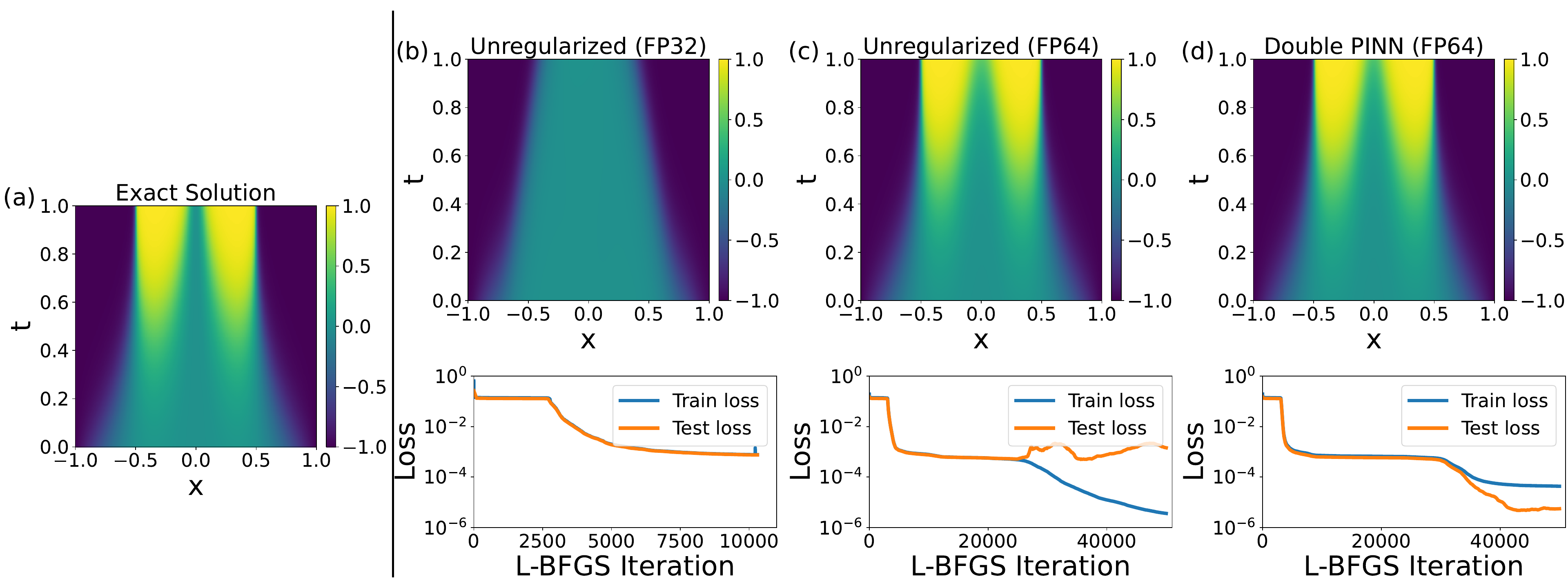}
    \caption{Comparison of the solution to the Allen-Cahn equation (a), with PINNs (b)-(d). Some training issues are caused by stiff training dynamics, rather than failure. This is indicated in (b), where the test and train loss track each other closely, yet optimization stalls due to the stiffness of the equation. An upgrade to FP64 (c) fixes this, but overfitting during later optimization reduces the accuracy. Combining double PINNs with FP64 in (d) achieves the best performance.}
    \label{fig:allen}
\end{figure}

Consider the Allen-Cahn equation (App.~\ref{app:allen}), which is sometimes considered a failure mode equation. We push back on this characterization, and instead suggest that optimization dynamics cause large error without hitting a failure mode.
Figure~\ref{fig:allen} shows numerical solution of the Allen-Cahn equation in (a), along with a PINNs training run at FP32 (b). Here, we use a denser $4096$ point grid, in order to be able to capture the sharp interfaces in the solution. The FP32 PINN does not optimize well. Yet examining train and test losses show they move in lockstep, completely different behaviour from the observed failure modes of the convection and wave equations. This suggests an optimization difficulty, rather than a classic failure mode. If this is true, an upgrade to FP64 should indeed rescue optimization, without the need for a regularizer. This is exactly the behaviour seen in (c), with the unregularized FP64 PINN successfully modelling the equation. However, a late-stage increase in the test loss suggests that after learning the solution the model starts to memorize the data as optimization continues. In this case, a regularizer should be beneficial, even in the absence of a failure mode. Figure~\ref{fig:allen}(d) shows this, with the best performance obtained through a combination of FP64 PINN and double PINN.

\subsection{Optimizing the Performance of Double PINN}

While we have shown that switching to higher precision is neither necessary nor sufficient for eliminating failure modes, it remains true that using higher precision will improve the accuracy of the PINN. We test double PINN at FP64 on the four benchmark equations: Convection, reaction, wave, and Allen-Cahn. We use a total number of $400$, $256$, $144$, and $4096$ grid domain points for each equation, respectively. The convection equation uses $200$ points for each boundary and initial condition, while the Reaction, Wave, and Allen-Cahn equations use $100$. Furthermore, we compare against the top three performing PINNs in the literature: unregularized vanilla FP64 PINN~\cite{xu2025fp64}, PINNsFormer~\cite{zhao2024pinnsformer}, and PINNMamba~\cite{xu2025sub}, all of which use a denser $10201$ grid. 

\begin{figure}[t]
    \centering
    \includegraphics[width=\linewidth]{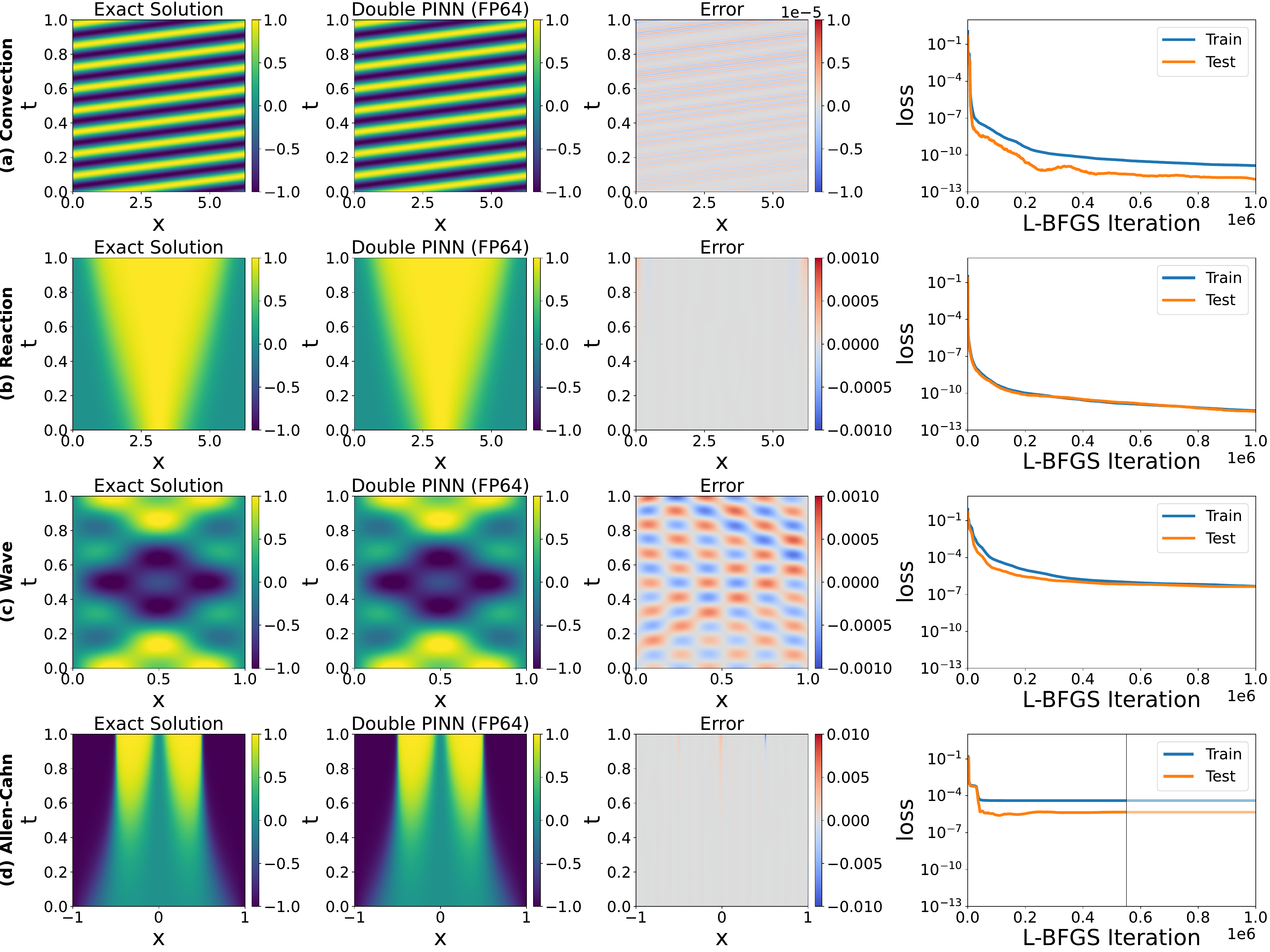}
    \caption{Regularization removes PINNs failure modes, while FP64 improves optimization. Combining both allows for a simple PINN architecture that achieves state of the art with few collocation points. Equations (a)-(c) trained for the full number of iterations, while (d) existed early after the train loss plateaued (final iteration indicated by the virtual line).}
    \label{fig:compare}
\end{figure}

\begin{table}
\centering
\caption{Comparison of double PINN at FP64 with the current top performing methods}
\label{tab:compare}
\small
\setlength{\tabcolsep}{4pt}
\begin{tabular}{l lllr}
\toprule
Model & \textsuperscript{3}Loss & rMAE & rRMSE & \textsuperscript{2}$N$ \\
\midrule
\multicolumn{5}{c}{Convection} \\
\cmidrule(lr){1-5}
\textsuperscript{1}PINNsFormer~\cite{zhao2024pinnsformer} & 9.0e-4 $\pm$ 1.0e-4 & 3.3e-2 $\pm$ 6.8e-3 & 4.4e-2 $\pm$ 7.3e-3 & 10403 \\
\textsuperscript{1}PINNMamba~\cite{xu2025sub} & 1.0e-4 $\pm$ 2.0e-5 & 1.8e-2 $\pm$ 3.7e-3 & 2.0e-2 $\pm$ 3.8e-3 & 10403 \\
PINN\_FP64~\cite{xu2025fp64} & 5.0e-6 $\pm$ 1.0e-6 & 5.9e-3 $\pm$ 1.3e-3 & 7.2e-3 $\pm$ 1.7e-3 & 10403 \\
\rowcolor{gray!15} double PINN (Ours) & 4.0e-11 $\pm$ 4.1e-11 & \textbf{4.9e-6 $\pm$ 4.4e-6} & \textbf{5.4e-6 $\pm$ 4.6e-6} & \textbf{800} \\
\midrule
\multicolumn{5}{c}{Reaction} \\
\cmidrule(lr){1-5}
\textsuperscript{1}PINNsFormer~\cite{zhao2024pinnsformer} & 3.0e-6 $\pm$ 1.0e-6 & 1.5e-2 $\pm$ 1.3e-3 & 3.0e-2 $\pm$ 2.7e-3 & 10403 \\
\textsuperscript{1}PINNMamba~\cite{xu2025sub} & 1.0e-6 $\pm$ 1.0e-6 & 9.2e-3 $\pm$ 1.7e-3 & 2.1e-2 $\pm$ 3.6e-3 & 10403 \\
PINN\_FP64~\cite{xu2025fp64} & 1.0e-5 $\pm$ 5.0e-6 & 2.7e-2 $\pm$ 6.3e-3 & 5.0e-2 $\pm$ 1.1e-2 & 10403 \\
\rowcolor{gray!15} double PINN (Ours) & 3.5e-12 $\pm$ 3.5e-13 & \textbf{1.1e-5 $\pm$ 2.3e-6} & \textbf{2.9e-5 $\pm$ 8.0e-6} & \textbf{456} \\
\midrule
\multicolumn{5}{c}{Wave} \\
\cmidrule(lr){1-5}
\textsuperscript{1}PINNsFormer~\cite{zhao2024pinnsformer} & 2.3e-2 $\pm$ 1.7e-3 & 3.5e-1 $\pm$ 8.7e-2 & 3.6e-1 $\pm$ 8.7e-2 & 10504 \\
\textsuperscript{1}PINNMamba~\cite{xu2025sub} & 2.0e-4 $\pm$ 3.0e-5 & 1.9e-2 $\pm$ 3.3e-3 & 2.0e-2 $\pm$ 3.3e-3 & 10504 \\
PINN\_FP64~\cite{xu2025fp64} & 4.2e-5 $\pm$ 1.6e-5 & 8.0e-3 $\pm$ 3.2e-3 & 8.1e-3 $\pm$ 3.1e-3 & 10504 \\
\rowcolor{gray!15} double PINN (Ours) & 3.5e-7 $\pm$ 8.9e-8 & \textbf{3.8e-4 $\pm$ 1.0e-4} & \textbf{3.8e-4 $\pm$ 1.0e-4} & \textbf{444} \\
\midrule
\multicolumn{5}{c}{Allen--Cahn} \\
\cmidrule(lr){1-5}
\textsuperscript{1}PINNsFormer~\cite{zhao2024pinnsformer} & 4.6e-1 $\pm$ 2.9e-1 & 9.9e-1 $\pm$ 4.0e-2 & 9.9e-1 $\pm$ 4.2e-2 & 10504 \\
\textsuperscript{1}PINNMamba~\cite{xu2025sub} & 2.7e-3 $\pm$ 2.0e-4 & 1.4e-1 $\pm$ 1.2e-2 & 2.7e-1 $\pm$ 2.0e-2 & 10504 \\
PINN\_FP64~\cite{xu2025fp64} & 1.3e-5 $\pm$ 4.0e-6 & 1.6e-2 $\pm$ 3.6e-3 & 5.5e-2 $\pm$ 1.1e-2 & 10504 \\
\rowcolor{gray!15} double PINN (Ours) & 4.0e-5 $\pm$ 6.1e-10 & \textbf{9.6e-5 $\pm$ 2.0e-5} & \textbf{3.6e-4 $\pm$ 9.8e-5} & \textbf{4396} \\
\bottomrule

\multicolumn{5}{l}{\tiny \textsuperscript{1} PINNsFormer and PINNMamba data, and high-resolution numerical solution for Allen-Cahn taken from Ref.~\cite{xu2025fp64}} \\
\multicolumn{5}{l}{\tiny \textsuperscript{2} $N$ is the total number of collocation points (domain + initial + boundary).} \\
\multicolumn{5}{l}{\tiny \textsuperscript{3} Loss is not directly comparable across the methods due to the regularization penalty, but is included here for reference.}

\end{tabular}
\end{table}

Figure~\ref{fig:compare} shows the result of this training for $10^6$ L-BFGS iterations on the four benchmark equations, all of which achieve excellent error. The final loss values, relative Mean Absolute Error (rMAE), and relative Root Mean Square Error (rRMSE) are shown in Table~\ref{tab:compare}, along with the total number of collocation points $N$ used in each. Double PINN at FP64 performs better than the compared methods in both error metrics by at least an order of magnitude, achieving a new state-of-the-art. Note the Allen-Cahn requires significantly more points, as it developed sharp interfaces that are difficult to resolve.

\begin{figure}
    \centering
    \includegraphics[width=\linewidth]{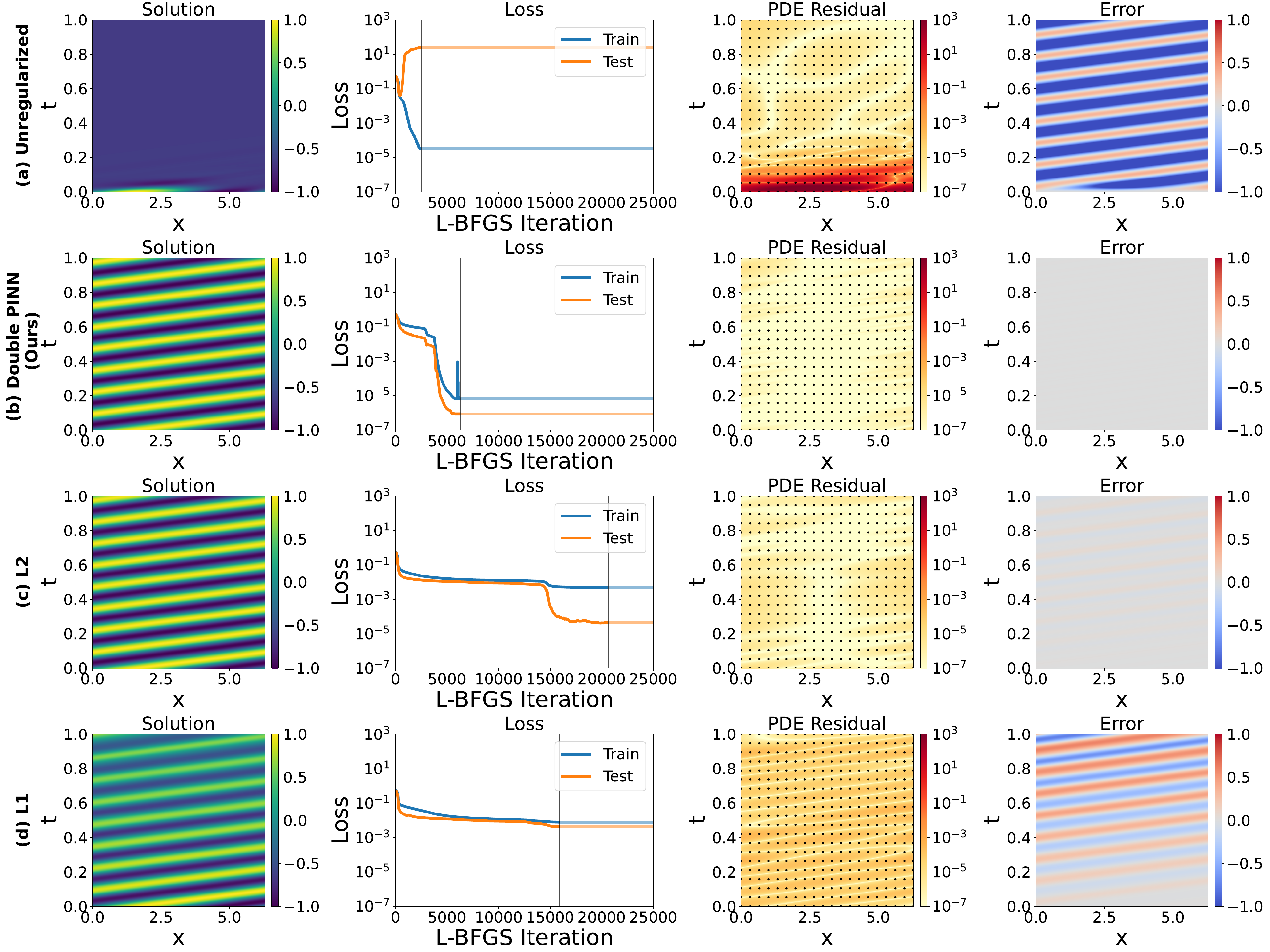}
    \caption{
    Comparison of vanilla PINNs on the $\beta=50$ convection equation at FP32. (a) unregularized PINNs fails due to overfitting of the residual. (b) Double PINNs succeed after a little over 5000 iterations of L-BFGS optimization. (c) $L^2$ regularized PINNs also succeeds, but requires more iterations, and maintains a higher test loss. (d) $L^1$ PINNs avoid failure from overfitting, but are difficult to optimize, resulting in large error. Final optimization steps are indiciated by the virtical line in the loss curves, and the final loss is indicated by the faded lines to enable easier visual comparison.}
    \label{fig:reg}
\end{figure}

Finally, a note should be given on the computational cost. Double backpropagation requires taking the gradients of the loss terms with respect to the inputs, which theoretically doubles the computational cost. This is significant. Yet, as shown in Figure~\ref{fig:reg}, double PINN is able to converge much faster as well. This, combined with the computational savings from using fewer collocation points, means that training of the regularized PINN becomes much cheaper in practice. This is especially true compared to architectures that use transformers or state space models, such as PINNsFormer and PINNMamba~\cite{zhao2024pinnsformer,xu2025sub}.

\section{Regularization of PINNs} \label{sec:reg}

Analysis in Sec.~\ref{sec:overfitting} points to overfitting being the underlying cause of PINNs failure modes, and that regularization with double PINN (Eq.~\eqref{eq:double_pinn}) can avoid such collapse while allowing the use of a vanilla PINN architecture with far fewer collocation points than normal. There are, however, several common choices of regularizers in the literature that might also be expected to work. This section discusses their performance when applied to PINNs in comparison to double PINN.

\subsection{Regularization Strategies}

We consider three regularization methods here. One common category are regularizers that shrink the size of the weights. Two of the most common of this type are $L^2$ and $L^1$ regularization, which penalize the norm of the combined sum of the weights. Specifically, if the weight parameters are denoted by $\mathbf{w}$, and the unregularized total loss as $\mathcal{L}_0$, then $L^2$ regularizer adds the $L^2$ norm of the weights as an auxiliary term, giving
$\mathcal{L}_\theta = \mathcal{L}_0 + (\lambda_r/2)\| \mathbf{w} \|^2_2$
where $\lambda_r$ controls the regularization strength~\cite{goodfellow2016deep}. The $L^1$ regularizer has the same form, swapping out the $L^2$ norm for the $L^1$ norm:
$\mathcal{L}_\theta = \mathcal{L}_0 + \lambda_r\| \mathbf{w} \|_1$.
Using an $L^1$ norm promotes sparsity in the network.
Our final and preferred method is double PINN, defined in Eq.~\eqref{eq:double_pinn}.

The performance of each of these regularizers is shown on the convection equation with $\beta=50$ in Figure~\ref{fig:reg}. The unregularized case shows failure caused by overfitting. All three of the regularization methods remove the overfitting, which can be seen in both the test loss tracking the train loss, and in the residual plots. double PINN performs the best, achieving the lowest loss and fastest convergence, while the $L^1$ regularization performs the worst. We suspect that the sparsity enforced by $L^1$ reduces the expressiveness of the network too much. The $L^2$ regularizer enforces no such sparsity, and overcomes the failure mode. 
We attribute the better performance of double PINN to the fact that it does not limit the network weights directly, but rather acts only on the residual and so does not limit the expressivity of the network.

\section{Conclusions}

Failure modes in PINNs have received considerable attention over the past several years, with research attempting to determine both their underlying causes and effective solutions. Here we have reanalyzed the problem, and shown that the underlying cause is overfitting. This changes the current understanding of failure modes, and suggests regularization as the straightforward prescription. We show that without it, even systems that aim to solve the failure modes (such as high precision PINNs), implicitly rely on overdetermining the problem, adding significant computational cost in the form of a large number of collocation points.
By extending double-backpropagation to PINNs, we show that we can train efficiently, and with top tier performance, using a fraction of the points.
We expect the advances here will have broad benefits in PINNs, particularly on complex and high dimensional PDEs.

\begin{ack}

This work was partially supported by JST CREST (JPMJCR24Q5), ASPIRE (JPMJAP2329), and PRESTO (JPMJPR24TB), and JSPS KAKENHI (24K15105), Japan.

\end{ack}

{\small
\bibliography{ref.bib}
\bibliographystyle{unsrtnat}
}

\newpage

\appendix

\section{Experimental Setup} \label{app:exp}

Here the experimental setup is discussed, along with the training procedure. For the most part, we follow the standard vanilla description of PINNs~\cite{raissi2019physics}. 

\subsection{Network Setup}
Our implementation is written in JAX, using the Flax NNX library for the base network, the Optax library for the L-BFGS implementation, and the Orbax library for saving the model~\cite{jax2018github, flax2020github, deepmind2020jax}.
The network itself contains 4 hidden layers, each with 128 neurons. We use a hyperbolic tangent activation, and initialize the weights using the Glorot normal initializer~\cite{glorot2010understanding}. Inputs to the network are normalized from $[-1,1]$, matching the original vanilla PINNs implementation.

The defaults for Optax's L-BFGS optimizer vary slightly from those of the popular Pytorch version, with PyTorch's optimizer containing a notably higher memory setting of 100, instead of Optax's 10~\cite{deepmind2020jax, Ansel_PyTorch_2_Faster_2024}. We increase the memory to 100 to match the Pytorch implementation, but leave the rest of Optax's defaults alone.

The L-BFGS optimizer also contains no inherent stopping criteria. In our training algorithm, we therefore take the following approach: training proceeds with the relative difference in the current train loss noted every 100 iterations. We then compare it against the square root of the machine epsilon for whatever precision we were using on the run (either 32 bit floating point or 64 bit floating point). Taking the root of machine epsilon is standard in numerical optimization, and we find our method gives a good indication to when the optimizer is no longer making progress~\cite{press1992numerical}. We use the difference over 100 runs to avoid pre-mature exiting while the optimizer is traversing the loss landscape. Together, this means we optimize as long as there is a relative difference in the loss of $3.453 \times 10^{-4}$ at FP32, and $1.490 \times 10^{-8}$ at FP64 over 100 iterations. Finally, we also specify a maximum number of epochs. This is relevant mostly for floating point 64, as the increased precision allows the optimizer to continuously make improvements for many iterations without stalling.

We use two sets of collocation points during training: a training and a test set. The training set consists of a small grid of collocation points. For example, Fig.~\ref{fig:convec} uses $200$ points for each of the boundaries, and $400$ points as a grid over the domain. The test set consists of a much larger set of random points, meant to sample the solution over a larger area to indicate its broader performance without relying on knowing the analytical solution. This test set uses $10000$ points randomly selected over the domain, and $10000$ linearly spaced points for each of the boundary and initial conditions. We expect the train and test set to move together for a well-trained model, while a lower train loss compared to the test loss indicates overfitting. A divergence between the two indicates a catastrophic failure, where the loss is minimized only at the collocation points, at the expense of the solution elsewhere. We consider this divergence the hallmark of a PINNs failure mode.

\section{Partial Differential Equations}

The following section contains the problem formulations for each of the partial differential equations described in the main work.

\subsection{1D Convection or Linear Advection} \label{app:convection}
The 1D convection equation (sometimes called linear advection) is
\begin{equation}
    \frac{\partial u}{\partial t} + \beta \frac{\partial u}{\partial x} = 0,
    \quad x \in [0,2\pi], \ t \in [0,1]
\end{equation}
with initial and boundary conditions
\begin{gather}
    u(x,0) = \sin(x), \quad x \in [0,2\pi] \\
    u(0,t) = u(2\pi, t), \quad t \in [0,1]
\end{gather}
The analytic, closed form solution to the PDE is
\begin{equation}
    u_\text{exact} = \sin(x-\beta t)
\end{equation}
For the advection speed $\beta$, a common choice in the failure mode literature is $\beta =50$, which we use here.

\subsection{Wave Equation} \label{app:wave}
The 1D wave equation with Dirichlet boundary conditions is:
\begin{equation}
    \frac{\partial^2 u}{\partial t^2}
    - 4\frac{\partial^2 u}{\partial x^2} = 0,
    \quad x \in [0,1], \ t \in [0,1]
\end{equation}
with initial and boundary conditions
\begin{gather}
    u(x,0) = \sin(\pi x) + \frac{1}{2} \sin(\beta \pi x),
    \quad x \in [0,1] \\
    \frac{\partial u(x,0)}{\partial t} = 0,
    \quad x \in [0,1] \\
    u(0,t) = u(1,t) = 0, \quad t \in [0,1]
\end{gather}
where $\beta$ controls the frequency of the harmonic in the initial condition. We use $\beta=3$ here as is standard. The PDE has a closed form analytic solution
\begin{equation}
    u_\text{exact} = \sin(\pi x) \cos(2\pi t) 
    + \frac{1}{2} \sin(\beta \pi x) \cos(2\beta\pi t)
\end{equation}

\subsection{Reaction Equation} \label{app:reaction}
1D reaction equation with periodic boundary conditions:
\begin{equation}
    \frac{\partial u}{\partial t} - \rho u(1-u) = 0,
    \quad x \in [0,2\pi], \ t \in [0,1]
\end{equation}
with initial and boundary conditions
\begin{gather}
    u(x,0) = \exp\left(-\frac{(x-\pi)^2}{2(\pi/4)^2}\right),
    \quad x \in [0,2\pi] \\
    u(0,t) = u(2\pi, t), \quad t \in [0,1]
\end{gather}
The closed-form analytic solution is 
\begin{equation}
    u_\text{exact} 
    = \frac{\exp(-\frac{(x-\pi)^2}{2(\pi/4)^2}) \exp(\rho t)}
    {\exp(-\frac{(x-\pi)^2}{2(\pi/4)^2})(\exp(\rho t) -1) + 1}
\end{equation}
a common choice for the $\rho$, the growth-rate coefficient, is $\rho=5$, which we adopt here as well. As with the case of linear advection, the problem becomes more difficult for PINNs the larger the value of $\rho$.

\subsection{Allen-Cahn Equation} \label{app:allen}
1D Allen-Cahn equation with periodic boundary conditions:
\begin{equation}
    \frac{\partial u}{\partial t} - 0.0001 \frac{\partial^2 u}{\partial x^2}
    + 5u^3 - 5u = 0,
    \quad x \in [-1,1], \ t \in [0,1]
\end{equation}
with initial and boundary conditions
\begin{gather}
    u(x,0) = x^2 \cos(\pi x), \quad x \in [-1,1] \\
    u(-1,t) = u(1,t), \quad t \in [0,1] \\
    \frac{\partial u(-1,t)}{\partial x} 
    = \frac{\partial u(1,t)}{\partial x}, \quad t \in [0,1]
\end{gather}
The Allen-Cahn equation above is the only PDE we consider that does not have an exact solution. Instead, for comparison, we use a high-resolution solution from a spectral solver, obtained from the source code of Ref.~\cite{xu2025fp64}.

\section{Limitations} \label{app:limit}

Our method directly targets the mechanism of failure modes in PINNs. The $L^2$ and $L^1$ regularizers, while simple and cheap to implement, are not the strongest in terms of absolute performance and convergence speed. Our double backpropagation method, double PINN, drastically reduces the number of iterations needed for convergence, however requires a theoretical doubling of the computational cost due to the need to compute the extra gradients. This is made up for by the ability to use fewer points, and the more efficient training, but is still a noticeable limitation. While the method here removes failure modes, even when using single precision, it does not target the optimization procedure itself, meaning no benefit is gained on stiff equations that do not fail (though our method still improves late stage accuracy on these equations if they can be optimized effectively). Stiff equation and their complicated loss landscapes are different from failure modes, and have been discussed elsewhere in the literature.

\section{Broader Impacts} \label{app:broader}

Our method improves the training of PINNs by enabling failure-free training with few collocation points, and without needing to go beyond single precision neural networks. We expect this to be important for high-dimensional equations, which cannot be tackled with solvers that require dense collocation (the curse of dimensionality). Many modern graphics processing units are unoptimized for double-precision throughput, so being able to use single-precision is important for utilizing current hardware to its fullest. As PDE solving is important in many disciplines, e.g. engineering, the methods introduced here could have indirect broader impacts if used in those contexts. However, we do not feel the need to highlight anything specific beyond its direct application to PDE solving.

\end{document}